%% file: article.tex
\documentclass[twocolumn, USenglish]{article}

\usepackage{00_ps}
\input{00_acronyms}
\usepackage{scalefnt}
\usepackage{comment}
\usepackage{tikz}

\begin{document}

  \articletype{Tools}

  \author*[1]{Julius Beerwerth}
  \author[2]{Jianye Xu}
  \author[2]{Simon Schäfer}
  \author[2]{Fynn Belderink}
  \author[1]{Bassam Alrifaee}
  \runningauthor{J. Beerwerth et al.}
  \affil[1]{University of the Bundeswehr Munich}
  \affil[2]{RWTH Aachen University}
  \title{Zero-Shot MARL Benchmark in the Cyber-Physical Mobility Lab}
  \runningtitle{MARL Benchmark in the CPM Lab}
  
\abstract{
We present a reproducible benchmark for evaluating sim-to-real transfer of Multi-Agent Reinforcement Learning (MARL) policies for Connected and Automated Vehicles (CAVs). The platform, based on the Cyber-Physical Mobility Lab (CPM Lab) \cite{kloock2021cyberphysical}, integrates simulation, a high-fidelity digital twin, and a physical testbed, enabling structured zero-shot evaluation of MARL motion-planning policies. We demonstrate its use by deploying a SigmaRL-trained policy \cite{xu2024sigmarl} across all three domains, revealing two complementary sources of performance degradation: architectural differences between simulation and hardware control stacks, and the sim-to-real gap induced by increasing environmental realism. The open-source setup enables systematic analysis of sim-to-real challenges in MARL under realistic, reproducible conditions.}

  \journalname{at-Automatisierungstechnik}
  \journalyear{2026}
  \journalvolume{..}
  \journalissue{..}
  \startpage{1}
  \aop

\makeatletter
\renewcommand\ps@plain{%
  \renewcommand\@oddfoot{\hfil\parbox{\textwidth}{\vspace{10pt}\centering\footnotesize This is the Preprint of an article published by De Gruyter in at -- Automatisierungstechnik on May 7, 2026, available at \url{https://doi.org/10.1515/auto-2025-0057}.}\hfil}%
}
\makeatother
\thispagestyle{plain}

\maketitle

\section*{Open Material}

A CPM Lab testing guide will be published upon acceptance at: \url{https://bassamlab.github.io/cpm-lab-ros2/}

\section{Introduction}
\Ac{rl} has shown remarkable success in solving complex sequential decision-making problems by enabling agents to learn optimal behaviors through interactions with their environment. Building upon these advances, \ac{marl} has emerged as a powerful approach for motion planning in \acp{cav}, demonstrating strong performance in simulated environments \cite{hua2023marl}. The decentralized nature of \ac{marl} enables scalable coordination among multiple agents without the need for complex centralized control schemes, making it particularly appealing for real-world traffic \cite{oroojlooy2023review}.

Despite these advances, a major challenge remains: transferring policies trained in simulation to physical systems---a process commonly referred to as \emph{sim-to-real transfer} \cite{zhao2020simtoreal}. Although \ac{marl} agents often perform well in simulation, they frequently struggle when deployed on real hardware due to the \emph{sim-to-real gap}. This gap arises from discrepancies between simulation and the real world, such as sensor noise, actuation delays, model mismatches, and environmental variability. As a result, performance often degrades when deploying policies in physical systems without adaptation.

To bridge this gap, a variety of techniques have been proposed, including domain randomization \cite{tobin2017domain} and system identification \cite{kaspar2020sim2real}. However, most studies rely on simulation-based evaluation, lacking real-world deployments and reproducibility.

To address this limitation, we introduce a reproducible platform for systematically evaluating sim-to-real transfer in \ac{marl}-based motion planning. Our platform integrates three levels of realism: a simulation environment, a high-fidelity digital twin, and a physical testbed based on the \ac{cpmlab} \cite{kloock2021cyberphysical}. This setup enables structured, zero-shot deployment of \ac{marl} policies and facilitates direct comparison across simulation and real-world proxies.

\subsection{Related Work}

Various approaches have been proposed to bridge the sim-to-real gap, such as domain randomization, system identification, and data augmentation. Domain randomization involves introducing variability during training to improve the robustness of learned policies. Randomization strategies include 
\begin{enumerate*}[label=\arabic*)]
    \item observation randomization, such as injecting noise or varying lighting conditions \cite{tobin2017domain},
    \item physics randomization, including varying object properties like mass or friction \cite{andrychowicz2020learning}, and 
    \item dynamics randomization, such as perturbing actuation delays and control responses \cite{peng2018simtoreal}.
\end{enumerate*}
These techniques have been successfully applied to manipulation tasks, urban driving scenarios, and autonomous racing \cite{tobin2017domain,okelly2020f1tenth}. Beyond randomization, system identification aims to accurately calibrate simulated environments using real-world data \cite{kaspar2020sim2real}. Data augmentation improves policy robustness by diversifying visual and sensor inputs during training \cite{shu2021adversarial}. 

Most \ac{marl} studies remain confined to simulation due to the difficulty of deploying and synchronizing multiple physical agents. Small-scale testbeds provide a cost-efficient way to validate RL and \ac{marl} algorithms in real-world settings. Several testbeds have been developed to support such experiments, including Duckietown \cite{paull2017duckietown}, RoboRacer (formerly known as F1TENTH) \cite{okelly2020f1tenth}, and Robotarium \cite{wilson2021robotarium}. We refer to \cite{mokhtarian2024survey} for a comprehensive survey on small-scale testbeds. Our work is closely related to \cite{candela2022transferring}, which employs \ac{marl} for multiple \acp{cav} in the Duckietown environment. While \cite{candela2022transferring} investigates how domain randomization mitigates the sim-to-real gap in transferring \ac{marl} policies, we provide a reproducible benchmark platform for evaluating this sim-to-real gap. Furthermore, unlike \cite{candela2022transferring}, which uses a simple two-lane map, our \ac{cpmlab} map features an eight-lane intersection in the middle, a highway loop on the outside, and multiple on- and off-ramps to transition between them, as shown in \cref{fig:experiment-setup}.
While small-scale testbeds more faithfully represent a full-scale deployment compared to traditional computer simulations by introducing uncertainties commonly found in physical systems, they also have limitations. We analyzed the gap between small-scale and full-scale models in \cite{schafer2024smallscale}, showing that they excel in evaluating software components related to motion planning and communication but fall short in representing interactions with the surrounding environment.


Our prior work introduced \textit{SigmaRL} \cite{xu2024sigmarl}, a sample-efficient and generalizable \ac{marl} framework for motion planning of \acp{cav}. Generalization in \ac{rl} refers to the ability of a learned policy to perform well in unseen or varying environments that differ from those encountered during training. It directly affects sim-to-real transfer, because a policy must generalize from the simulated environment to the real world. We provide the details regarding the observation design in \cref{sec:observation-sigmarl}.

\subsection{Contribution of this Article}
\label{sec:contribution}

This article introduces a reproducible platform for evaluating sim-to-real transfer in \ac{marl}-based motion planning for \acp{cav}. We demonstrate its utility by deploying a policy trained with our previously proposed SigmaRL framework across three levels of realism: a simulation environment, a high-fidelity digital twin of the \ac{cpmlab}, and a physical deployment in the \ac{cpmlab}. The main contributions of this article are as follows:
\begin{enumerate}
	\item We present a reproducible platform for studying sim-to-real transfer in \ac{marl}-based motion planning.
	\item We provide a baseline evaluation using SigmaRL-trained policies across simulation, digital twin, and physical deployment.
	\item We release tools, configurations, and metrics for replicating experiments, and provide a systematic discussion of sim-to-real effects arising from domain shift and increasing system realism.
\end{enumerate}


\subsection{Organization of this Article}
\label{sec:organization}

This article is organized as follows. \Cref{sec:overview-sigmarl} introduces the SigmaRL framework and explains its approach to structured observation design for \ac{marl}. \Cref{sec:cpmlab} presents the \ac{cpmlab}, the experimental platform used for validation, and outlines its integration with the SigmaRL pipeline. \Cref{sec:eval} details the training procedure, experimental setup, and the evaluation results from the sim-to-real experiments. Finally, \cref{sec:conclusion} concludes the paper and outlines future research directions.


\section{Overview of SigmaRL} \label{sec:overview-sigmarl}
Our SigmaRL \cite{xu2024sigmarl} leverages Multi-Agent Proximal
Policy Optimization (MAPPO) \cite{yu2022surprising} and trains agents using decentralized actors (i.e., the policies) and a centralized critic. It models agents with the nonlinear kinematic bicycle model \cite{rajamani2006vehicle}. We briefly introduce the observation, reward, and action design of SigmaRL in \cref{sec:observation-sigmarl}, \cref{sec:reward-sigmarl}, and \cref{sec:action-sigmarl}, respectively. Further details can be found in our previous work \cite{xu2024sigmarl}.

\subsection{Observation Design} \label{sec:observation-sigmarl}
The key feature of SigmaRL is its efficient observation design. We integrate domain knowledge directly into the observation space and use structured observations that encode geometric and traffic-specific features, such as distances to road boundaries and lane centerlines. These features are designed to reflect task-relevant invariances, i.e., properties of the environment that remain important across a wide range of driving tasks and scenarios. For example, the distance to road boundaries is always relevant for lane keeping, regardless of the specific driving task or traffic context. By incorporating such invariant features directly into the observation space, we improve generalization and zero-shot transfer across diverse traffic scenarios. 

Let $\mathcal{N} \coloneqq \{1,\ldots,n_{\mathrm{agent}}\}$ denote the set of agents' indices, where $n_{\mathrm{agent}}$ denotes the number of agents. We will refer to agents by their indices.
At each discrete time step $t$, the observation of each agent $i \in \mathcal{N}$ is represented in an ego-centric coordinate system, i.e., the ego view of agent $i$. To obtain this observation for each agent $i$, we assume the states of all agents $\{\bm{s}_{t,\mathrm{SigmaRL}}^{(k)}\}_{k=1}^{n_{\mathrm{agent}}}$ in the global coordinate system are available. Let each agent $i$'s state at time step $t$ in the global coordinate system in SigmaRL be denoted as
\begin{equation}
    \bm{s}_{t,\mathrm{SigmaRL}}^{(i)} = \left[\bm{p}_{t}^{(i)}, \psi_{t}^{(i)}, \bm{v}_{t}^{(i)},  \sigma_{t}^{(i)} \right],
    \label{eq:internal-state-sigmal-rl}
\end{equation}
where $\bm{p}_{t}^{(i)} \in \mathbb{R}^2$ is the position, $\psi_{t}^{(i)} \in [-\pi, \pi]$ the orientation (yaw angle), $\bm{v}_{t}^{(i)} \in \mathbb{R}^2$ the velocities, and $\sigma_{t}^{(i)} \in [-\sigma_{\max}, \sigma_{\max}]$ the steering angle, with $\sigma_{\max}$ being the maximum steering angle. Furthermore, we assume \iac{hd} map is available, within which agents can localize themselves. Based on the global states of all agents and the \ac{hd} map data, we can easily compute the observation for each agent $i$ on its ego-centric coordinate system. If this data is not available, agents must be equipped with onboard sensors that allow them to observe quantities such as distances to road boundaries and surrounding agents, and the geometric shapes of these surrounding agents. In \cref{sec:cpmlab-observation}, we explain how we obtain the required observation \eqref{eq:internal-state-sigmal-rl} for each agent $i$ in our \ac{cpmlab}.

\subsection{Reward Design} \label{sec:reward-sigmarl}
The reward function $r_t^{(i)}$ for each agent $i$ consists of three components:
$$
r_t^{(i)} = r_{\mathrm{track}}^{(i)} + r_{\mathrm{speed}}^{(i)} + r_{\mathrm{penalty}}^{(i)}.
$$
The term $r_{\mathrm{track}}^{(i)}$ encourages the agent to follow its reference path (the road centerline), while $r_{\mathrm{speed}}^{(i)}$ rewards maintaining a higher forward speed to improve traffic efficiency. The term $r_{\mathrm{penalty}}^{(i)}$ imposes penalties for unsafe behavior, including collisions with other agents, collisions with road boundaries, and violating safety distances to nearby agents or boundaries. This formulation incentivizes agents to generate trajectories that are safe, smooth, and efficient in multi-agent scenarios.

\subsection{Action Design} \label{sec:action-sigmarl}
At each time step $t$, each policy $i$ produces a continuous action vector $\bm{u}_t^{(i)} = [u_{v,t}^{(i)}, u_{\delta,t}^{(i)}] \in \mathbb{R}^2$, where $u_{v,t}^{(i)} \in \mathbb{R}$ and $u_{\delta,t}^{(i)} \in \mathbb{R}$ denote the commanded speed and steering angle, respectively. To ensure the outputs are physically feasible, each component is constrained within task-specific limits by applying a squashing function to the sampled action. Specifically, the policy models a Gaussian distribution and passes the sampled action through a hyperbolic tangent (\texttt{tanh}) function, which bounds the output to $(-1, 1)$; this is then rescaled to the actual control range. This approach maintains differentiability and produces smooth and bounded control commands.


\section{Cyber-Physical Mobility Lab}
\label{sec:cpmlab}

In this chapter, we introduce the CPM Lab, our small-scale testbed and primary evaluation platform for this work. First, we provide an overview of the CPM Lab's setup and synchronization structure. Then, we detail how we mapped the CPM Lab’s state information to SigmaRL’s observation format and converted SigmaRL’s policy outputs into reference trajectories for the agents.

\subsection{Overview}

The CPM Lab \cite{kloock2021cyberphysical} is a small-scale testbed designed for evaluating \ac{cav} systems. It features 20 small-scale vehicles \cite{scheffe2020networked}, called {\textmu}Cars. An overhead ceiling camera \cite{kloock2020visionbased} provides absolute position estimates for every agent, and a synchronization layer manages the setup.

The CPM Lab's control architecture for the {\textmu}Cars is divided into three levels. At the lowest level, each {\textmu}Car's hardware is managed by a controller for speed and steering angle. The vehicle controller, a level above the low-level controller, continuously determines the current speed and steering angle required to follow a reference trajectory. Finally, at the topmost level, we have the motion planner, which logically integrates into the car's control stack but needs to be offboarded due to limited computational resources available on the {\textmu}Cars.
This planner computes a reference trajectory for all {\textmu}Cars while accounting for each agent's position and possible road boundary constraints.

A synchronization layer orchestrates the interaction between the motion planner and the CPM Lab. Whenever a reference update is needed, this layer ``ticks,'' prompting the motion planner to compute a new reference trajectory. Subsequently, the layer broadcasts the commands to the {\textmu}Cars, which apply them via their respective vehicle controllers. As part of each ``tick,'' the synchronization layer also provides the planner with an up-to-date state list of all agents.

When using SigmaRL, as a motion planner, two transformations are required. First, the state list from the CPM Lab must be adapted into SigmaRL's internal state format, and second, SigmaRL's policy output must be converted into a reference trajectory. The following subsections provide these two transformations.

\subsection{Observation} \label{sec:cpmlab-observation}

The CPM Lab provides a state list consisting of $n_{\mathrm{agent}}$ agents, each agent $i$ described by the state vector
\begin{equation}
    \bm{s}_{t,\mathrm{CPM}}^{(i)} = \left[\bm{p}_{t}^{(i)}, \psi_{t}^{(i)}, v_{t}^{(i)}, {\widetilde{\sigma}}_{t}^{(i)} \right],
    \label{eq:observer-cpm-lab}
\end{equation}
where $\bm{p}_{t}^{(i)} \in \mathbb{R}^2$ is the position, $\psi_{t}^{(i)} \in [-\pi, \pi]$ the orientation (yaw angle), $v_{t}^{(i)} \in \mathbb{R}$ the speed, and $\widetilde{\sigma}_{t}^{(i)} \in [-1, 1]$ the normalized steering angle. Because this format does not match SigmaRL’s required state representation \eqref{eq:internal-state-sigmal-rl}, we introduce a mapping function
\begin{equation*}
    f: \bm{s}_{t,\mathrm{CPM}}^{(i)} \mapsto  \bm{s}_{t,\mathrm{SigmaRL}}^{(i)}.
\end{equation*}

Under this mapping, the position $\bm{p}_{t}^{(i)}$ and orientation $\psi_{t}^{(i)}$ remain unchanged, and the steering angle $\widetilde{\sigma}_{t}^{(i)}$ is denormalized via
\begin{equation*}
    \sigma_{t}^{(i)} = \sigma_{\max}  \widetilde{\sigma}_{t}^{(i)}
\end{equation*}
where $\sigma_{\max}$ corresponds to the {\textmu}Car’s maximum steering angle. Next, the speed $v_{t}^{(i)}$ is converted to a 2D velocity vector $\bm{v}_{t}^{(i)}$ by assuming the kinematic bicycle model (with the center of gravity and geometric center coinciding):
\begin{equation*}
    \bm{v}_{t}^{(i)} = v_{t}^{(i)} \begin{bmatrix}\cos \beta_{t}^{(i)} \\\sin \beta_{t}^{(i)} \end{bmatrix},
\end{equation*}
where
\begin{equation*}
    \beta_{t}^{(i)} = \arctan \left(\frac{\ell_r}{\ell_{wb}} \tan\sigma_{t}^{(i)} \right).
\end{equation*}

Here, $\ell_{wb}$ denotes the wheelbase and $\ell_r$ the rear wheelbase.

\subsection{Action}

In the CPM Lab, several control schemes are available to control the agents. The most suitable mode is the trajectory-following mode, which requires converting SigmaRL’s outputs into a continuous reference trajectory.  

In a typical \ac{mpc} setup, two horizons are considered: the control horizon $H_c$ and the prediction horizon $H_p$. The control horizon $H_c$ specifies how many steps ahead new control input are generated, while the prediction horizon $H_p$ limits the step at which we use a motion model to predict the future states, enabling closer adherence to a reference path. In this work, we adapt this control schema to integrate an RL-based motion planner with the CPM Lab’s trajectory-following capabilities. Rather than performing the optimization phase of a typical \ac{mpc} loop, we rely on the RL policy outputs to generate the trajectory.

During the control horizon $H_c$, actions provided by the RL policy dictate required control inputs $\bm{u}_{{t+j}}^{(i)}$. These inputs are then combined with the kinematic bicycle model to predict the subsequent states $\bm{s}_{t,\mathrm{SigmaRL}}^{(i)}$. This predicted state serves as a basis for making further policy decisions at predicted time $t + k$, with this process looping until reaching $H_c$. Once we reach $H_c$ at time $t + H_c\), we transition our approach: we maintain constant velocity based on the final action derived from the policy while gradually reducing steering input $|u_\sigma|$ to zero over the remaining prediction horizon.

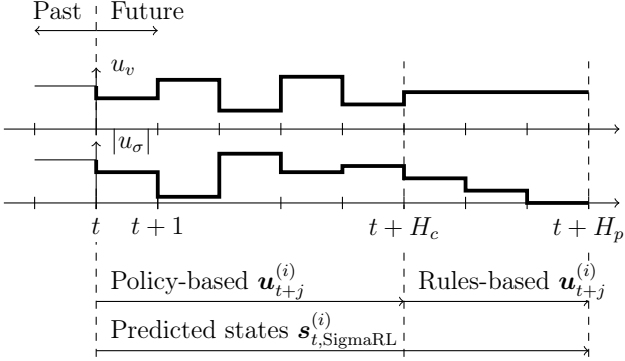
\begin{figure}[tb]
    \centering
    \input{figures/prediction-illustration}
    \caption{Illustration of how SigmaRL's policy output is applied within the control horizon and then transitioned to a rules-based approach for the remaining prediction horizon, yielding a complete trajectory for each agent $i$ in the scene.}
    \label{fig:test-illustration}
\end{figure}

\Cref{fig:test-illustration} illustrates how the velocity and steering-angle inputs, labeled $u_v$ and $u_\sigma$, evolve over time. From time $t$ to $t + H_c$, actions computed by the RL policy are applied. After reaching time $t + H_c$, we hold constant our last derived action for velocity $u_v$ while systematically diminishing steering input $|u_\sigma|$ until it reaches zero. Throughout each timestep within both horizons, our motion model updates predicted states accordingly.

\section{Evaluation}
\label{sec:eval}

We present a baseline evaluation to assess the zero-shot sim-to-real transferability of a policy trained in simulation using SigmaRL. Code reproducing the simulation results in SigmaRL is available\footnote{\url{https://github.com/bassamlab/SigmaRL}}.

\subsection{Training in SigmaRL} \label{sec:training-sigmarl}
\begin{figure*}[!tb]
  \centering
  \input{figures/experiment-overview}
  \caption{Example initial setup and representative trajectories for one configuration.
One representative run per environment is shown, using the same initial positions across all environments; the run was selected from the digital twin as the one with centerline deviation closest to that environment's mean.
Trajectories are shown over the \SI{18}{\second} evaluation horizon; in the physical lab, a collision during this interval can further truncate the trajectory.
Higher point density near the intersection occurs because vehicles slow down or briefly stop, and positions are sampled at a fixed rate.
(a) Initial setup with assigned reference paths. (b) Trajectories in the SigmaRL simulation.
(c) Trajectories in the digital twin. (d) Trajectories in the physical CPM Lab.}
  \label{fig:experiment-setup}
\end{figure*}
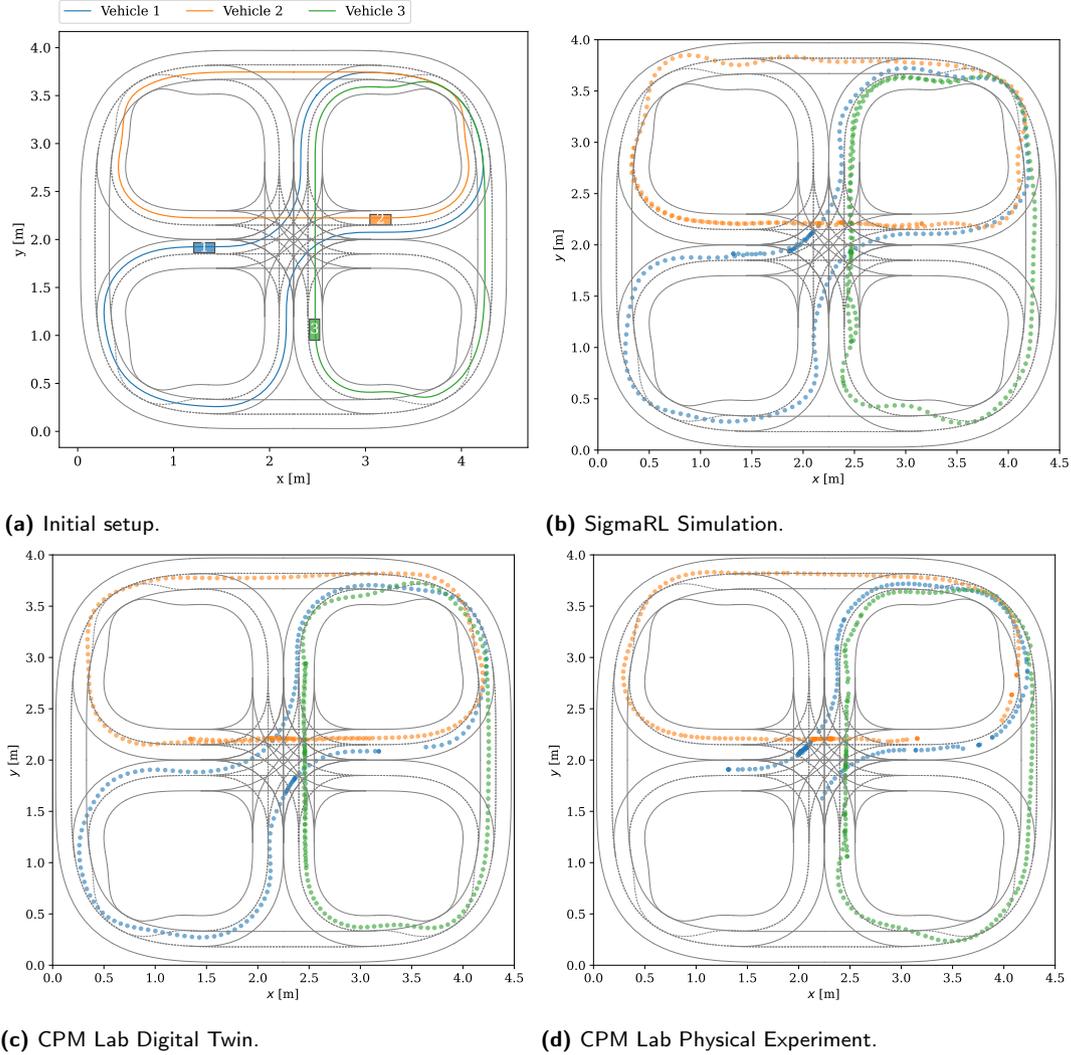

SigmaRL provides a variety of traffic scenarios, one of which resembles the traffic environment of the \ac{cpmlab}. To evaluate the generalization of SigmaRL, we trained a policy solely on the intersection segment of the map shown in \cref{fig:experiment-setup} and tested it on the entire map. We used $n_\mathrm{agent}=4$ agents for training, with randomly initialized states and reference paths. Once an agent collided or exited the intersection area, we reset it with a new random configuration.

For both the actor and the critic, we used a simple feedforward neural network with three hidden layers of 
size 256 and \texttt{tanh} activation functions. The observation size in our SigmaRL is 32, which serves as the input vector for the actor. Consequently, the total number of trainable parameters becomes $(32\cdot256 + 256) + (256\cdot256 + 256) + (256\cdot256 + 256) + (256\cdot4 + 4) \approx 140k$, making the model lightweight (smaller than \SI{0.5}{\mega\byte} in size) and suitable for real-time deployment. We directly transferred the trained actor to physical vehicles in the \ac{cpmlab} without any additional fine-tuning.

We trained the policy for 1000 iterations, collecting a batch of $2^{12}$ samples in each iteration, resulting in approximately four million samples in total. To improve training efficiency, we adopted minibatch training with a minibatch size of $2^9$, and reused the collected samples by training for 30 epochs per iteration. After each iteration, we retained the best-performing policy during the whole training process.


\begin{table}[tb]
    \caption{Parameters used in simulations and physical experiments.}
    \centering
    \begin{tabular}{ll}
        Parameter & Value \\
        \midrule
        Vehicle length $\ell$, width $w$ & \SI{0.22}{\meter}, \SI{0.107}{\meter} \\
        Wheelbase $\ell_{wb}$, rear wheelbase $\ell_r$ & \SI{0.15}{\meter}, \SI{0.075}{\meter} \\
        Max. steering angle & \SI{31}{\degree}\\
        Max. steering rate & \SI{90}{\degree\per\second}\\
        Max. (min.) acceleration & \SI{5}{\meter\per\second\squared} (\SI{-5}{\meter\per\second\squared}) \\
    \end{tabular}
    \label{tab:parameters}
\end{table}

\subsection{Experimental Setup}
\label{sec:experiment-setup}

To evaluate the transferability of our policy, we conduct zero-shot deployment across three environments: simulation, a high-fidelity digital twin, and the physical \ac{cpmlab}. These environments differ along several axes, including dynamics modeling, localization accuracy, actuation delay, and execution infrastructure. 

The simulation environment uses a kinematic bicycle model with instantaneous, noise-free actuation, where policy actions are applied directly to the vehicle model. In contrast, the digital twin uses a grey-box vehicle model identified from real-world logs and includes sensor and actuation imperfections; policy outputs are converted into reference trajectories that are tracked by a mid-level trajectory follower implemented as an MPC. The physical \ac{cpmlab} uses scaled vehicles, each subject to motor delays, sensor noise, and imperfect localization via an overhead camera system with known latency~\cite{kloock2020visionbased}, and applies the same mid-level trajectory follower (MPC) as the digital twin. As a 1:18-scale testbed, the CPM Lab does not fully replicate full-scale vehicle dynamics. Typical operating speeds are below $\SI{2}{\meter\per\second}$, and geometric scaling is not proportional to dynamic scaling, resulting in different relative effects of inertia, friction, and actuation delays.

To ensure statistical robustness, each of the three trained policies was evaluated in all environments for three different initial position configurations, with three repetitions per configuration. This yields $3 \cdot 3 \cdot 3 = 27$ runs per environment. For paired comparison, the initial positions in the simulation and digital twin were matched to those measured in the corresponding physical lab runs. We emphasize that reproducibility in the CPM Lab refers to deterministic computation and synchronization, enabling fair repeated trials. Nonetheless, measurement noise, actuation uncertainties, and stochastic policies naturally lead to variation across runs, which we capture through statistical evaluation.

We use $n_\text{agent}=3$ agents for the evaluation. Although the CPM Lab can operate a larger number of vehicles, preliminary physical experiments with higher agent counts (e.g., five agents) resulted in frequent collisions and required repeated manual interventions; we therefore restrict the evaluation to three agents to ensure stable and reproducible trials. Each run lasted \SI{18}{\second}, corresponding to 180 steps at a fixed timestep of \SI{0.1}{\second}. In the physical experiment, agents were manually reset following collisions. Each agent was repositioned near its last valid location and heading.

\subsection{Performance Metrics}
\label{sec:metrics}

To quantify safety and performance across environments, we evaluate four metrics, each normalized to the total distance traveled to enable fair comparison between runs. Unless otherwise stated, values are reported per \SI{100}{\meter} of aggregate agent travel. Metrics are computed per run and we report mean ± std and IQM across all runs per environment.

\textbf{Agent--Agent Collision Rate (CRA-A):}  
A collision event is detected based on the minimum translation vector between two agent geometries, which becomes negative when agents overlap. To reduce false positives from localization noise, we apply a hysteresis filter: an overlap must persist for at least~3 consecutive steps to start a collision, and at least~5 consecutive non-colliding steps are required to end it. The CRA-A value is the number of detected events divided by the total distance traveled, reported in events per \SI{100}{\meter}.

\textbf{Agent–Lane Collision Rate (CRA-L):}  
To measure lane keeping, we use polygon intersection between the agent’s footprint and the lane boundaries. 
A violation is recorded for any timestep where the footprint intersects or lies outside the lane polygon. 
To avoid counting negligible boundary grazes due to tight lane definitions or localization noise, a \SI{10}{\percent} slack relative to the vehicle width is applied, 
so that small deviations within this margin are ignored. 
For each run, the violation length is the sum of the distances traveled while outside the lane (beyond the slack), normalized by the total distance traveled, and reported in meters per \SI{100}{\meter}. Thus, if an agent has a CRA-L of \SI{40}{\meter\per \SI{100}{\meter}}, the agent violated the lane boundaries for \SI{40}{\meter} of the \SI{100}{\meter} traveled.

\textbf{Centerline Deviation (CD):}  
The mean lateral distance between the agent’s center and its reference path (road centerline), computed over the entire run. This evaluates tracking accuracy and penalizes systematic drift. It is reported in meters.

\textbf{Average Speed (AS):}  
The mean forward speed of each agent, computed as the $\ell^2$~norm of the velocity vector. This metric reflects traffic efficiency and indirectly captures hesitation or overly conservative behavior in response to noise or unexpected disturbances. It is reported in \si{\meter\per\second}.

\begin{table*}[t]
\caption{Mean $\pm$ standard deviation across $n{=}27$ runs per environment.
IQM values are shown in parentheses. Arrows indicate the direction of improvement.
CRA-A is the agent–agent collision rate per \SI{100}{\meter};
CRA-L is the distance traveled outside the lane per \SI{100}{\meter};
CD is the mean centerline deviation; AS is the mean speed.
The SigmaRL simulation applies policy actions directly, whereas the digital twin and physical lab use a mid-level MPC trajectory follower.}
\label{tab:env_summary_compact}
\centering
\small
\begin{tabular}{lcccc}
\toprule
Environment
& {CRA-A $\downarrow$ [\si{events\per \SI{100}{\meter}}]}
& {CRA-L $\downarrow$ [\si{\meter\per \SI{100}{\meter}}]}
& {CD $\downarrow$ [\si{\meter}]}
& {AS $\uparrow$ [\si{\meter\per\second}]} \\
\midrule
SigmaRL Simulation (no MPC)
& \textbf{0.37} $\pm$ 1.33 (0.47)
& 45.99 $\pm$ 6.69 (47.12)
& 0.050 $\pm$ 0.009 (0.050)
& 0.759 $\pm$ 0.024 (0.761) \\
Digital Twin (with MPC)
& 2.10 $\pm$ 2.80 (1.47)
& \textbf{38.80} $\pm$ 9.57 (38.40)
& \textbf{0.041} $\pm$ 0.010 (0.039)
& \textbf{0.771} $\pm$ 0.031 (0.777) \\
Physical Lab (with MPC)
& 4.49 $\pm$ 6.26 (2.45)
& 41.09 $\pm$ 7.21 (41.10)
& 0.044 $\pm$ 0.010 (0.043)
& 0.728 $\pm$ 0.035 (0.750) \\
\bottomrule
\end{tabular}
\end{table*}

\subsection{Results and Discussion}
\label{sec:results}

\Cref{tab:env_summary_compact} summarizes the performance metrics across the three deployment environments: simulation, high-fidelity digital twin, and the physical \ac{cpmlab}.  
All results are aggregated over $n{=}27$ runs per environment (three training seeds, three initial position configurations, and three repetitions each), with metrics reported as mean~$\pm$~standard deviation and interquartile mean (IQM) in parentheses.

The SigmaRL policies were trained solely on an intersection scenario and deployed without fine-tuning to a more complex evaluation setup including highway loops, on-ramps, and intersections. As a result, all reported metrics for SigmaRL already reflect a domain shift due to the mismatch between training and evaluation scenarios. Note that although the number of agents during training (four) differs from the number of vehicles during deployment (three), this difference does not significantly contribute to the domain shift. This is because the policies are decentralized with shared parameters, and each agent observes only a fixed number of nearby agents. As a result, the agent-level observation remains consistent regardless of the total number of agents in the environment. In addition, when comparing across environments (Simulation, Digital Twin, and Physical Lab), the results are further influenced by differences in vehicle dynamics, localization accuracy, and control behavior. We do not differentiate between the effects of domain shift and environment shift on the performance. However, our previous work on information-dense observation design \cite{xu2024sigmarl} showed that domain shift has small impact on performance, as the policies generalize well to unseen map regions. Since this domain shift is present across all environments, the remaining differences between the Digital Twin and the Physical Lab mainly capture the effects of environment shift. We attribute amplifications of domain-shift effects to the environment shift itself, as it is the underlying factor causing the observed degradation. Accordingly, this work does not attempt to experimentally disentangle domain shift from environment shift; instead, it focuses on comparing deployment environments under a fixed training domain, with a detailed ablation of domain shift left for future work.

Two main forms of performance degradation can be observed.  
Among all metrics, the number of collisions (CRA-A) serves as the most direct and robust indicator of overall performance, as it consistently increases across the deployment environments—from the SigmaRL simulation to the digital twin and the physical lab—while the other metrics are more sensitive to differences in architecture and control execution.

\paragraph*{Sim-to-real degradation}
The first and expected form of degradation appears in the transition from the digital twin to the physical lab, where all metrics deteriorate slightly.  
CRA-A increases from $2.10$ to $4.49$~events/100\,m, accompanied by a modest drop in average speed (AS) from \SI{0.771}{\meter\per\second} to \SI{0.728}{\meter\per\second}, and slightly higher CRA-L and CD values.  
These trends indicate a measurable sim-to-real gap arising from increased environmental realism, including sensor noise, actuation delays, and communication latency.  
As the environment becomes more physically accurate, the control task becomes more demanding, leading to a natural degradation in performance.  
This degradation pattern provides a consistent, quantitative measure of environment shift under an identical control architecture.

\paragraph*{Architectural degradation}
A second, less expected form of degradation emerges when comparing the SigmaRL simulation to the digital twin and the physical lab.  
While the SigmaRL simulation achieves the lowest CRA-A (0.37~events/100\,m), it shows worse performance in terms of lane violations (CRA-L), centerline deviation (CD), and average speed (AS).  
This behavior can be attributed to architectural differences in how the control commands are executed.  
In SigmaRL, the policy’s control outputs are applied directly to the vehicle dynamics model, whereas in the digital twin and the physical lab, the planned trajectories are passed to a mid-level \ac{mpc}.  
This \ac{mpc} smooths high-frequency control inputs and improves trajectory tracking, resulting in fewer short-lived boundary crossings and reduced lateral oscillations.  
Consequently, the superior CRA-L, CD, and AS observed in the digital twin and physical lab stem from this additional control layer rather than from differences in the learned policy itself. Integrating the CPM Lab’s mid-level \ac{mpc} into the SigmaRL training simulation would require a substantial redesign of the training environment; therefore, the simulation results reflect the original direct-control setup used during policy learning.

Overall, the results reveal two distinct yet complementary effects:  
(i)~a clear sim-to-real degradation from the digital twin to the physical lab, primarily captured by the rising number of collisions, and  
(ii)~an architectural performance difference between the direct-control setup in SigmaRL and the hierarchical control in the digital twin and physical lab.  
Together, these findings show that collision rate is the most consistent indicator of overall robustness across environments, whereas lane keeping, trajectory smoothness, and speed are more sensitive to control architecture and environment fidelity.

\section{Conclusion and Outlook}
\label{sec:conclusion}

This paper introduced a reproducible platform for evaluating the sim-to-real transferability of \ac{marl} policies for Connected and Automated Vehicles (CAVs).  
The benchmark spans three progressively realistic environments—simulation, digital twin, and a physical small-scale testbed—and enables structured zero-shot evaluation.

To demonstrate the benchmark’s capabilities, we deployed a policy trained with the SigmaRL framework across all three domains without fine-tuning.  
The evaluation revealed two complementary forms of performance degradation: (i)~a systematic increase in collisions from simulation to the digital twin and the physical lab, reflecting the growing realism and environmental disturbances; and (ii)~architectural differences between the direct-control setup in SigmaRL and the hierarchical control stack of the digital twin and physical lab, which influence secondary metrics such as lane keeping, trajectory smoothness, and average speed.  
Among all metrics, the number of collisions proved to be the most consistent and robust indicator of sim-to-real performance degradation. In contrast, metrics based on continuous deviations (CRA-L, CD, AS) are inherently more sensitive to evaluation horizon, trajectory smoothing, and control execution details, which explains why their absolute values vary more strongly across environments.

Despite these differences, the policies demonstrated strong resilience: agents recovered from collisions and disturbances, confirming the benefits of structured, semantically meaningful observations for generalization.  
Beyond these findings, this work highlights the practical value of small-scale physical testbeds for controlled and repeatable \ac{marl} experiments.  
While not a replacement for full-scale deployment, the \ac{cpmlab} provides a safe and scalable intermediate platform with the infrastructure needed for high-throughput sim-to-real validation.

In future work, we plan to relax the assumption of global state access and investigate partial observability, onboard perception, and communication-limited coordination.

\begin{acknowledgement}
This research was supported by the Bundesministerium für Verkehr (German Federal Ministry of Transport) within the project ``Harmonizing Mobility'' (grant number 19FS2035A) and by the Collaborative Research Center / Transregio 339 of the Deutsche Forschungsgemeinschaft (German Research Foundation).
\end{acknowledgement}

\bibliographystyle{IEEEtran}
\bibliography{00_bassam_publications, 00_literature}


\section*{Author Information}

\begin{minipage}{0.3\columnwidth}
\includegraphics[width=1in,height=1.25in,clip,keepaspectratio]{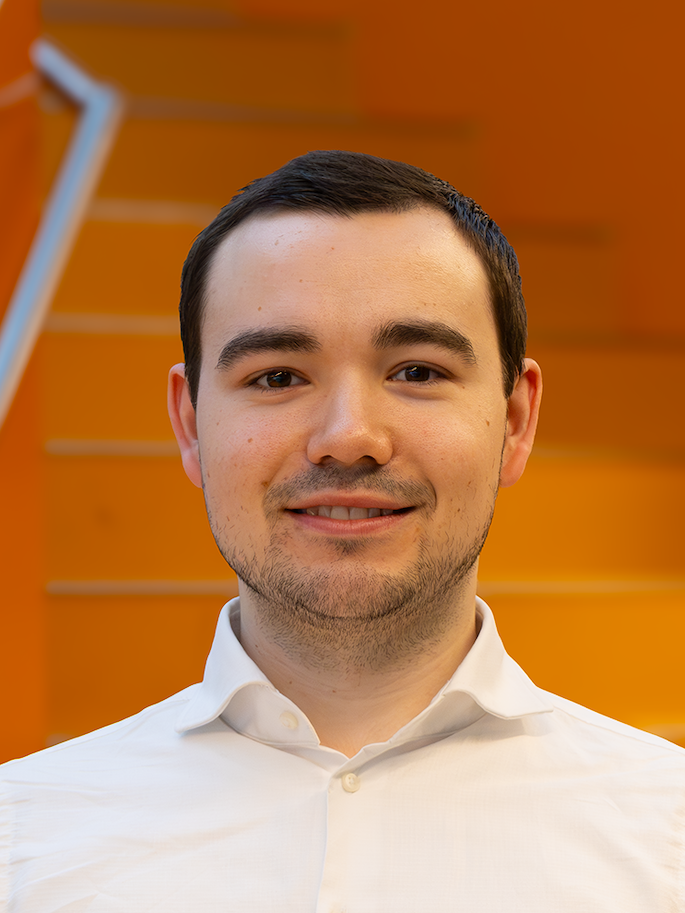}
\end{minipage}
\begin{minipage}{0.7\columnwidth}
\textbf{Julius Beerwerth} \\
Dept. of Aerospace Engineering, \\University of the Bundeswehr Munich, \\
85579 Neubiberg, \\
Germany \\
\textbf{julius.beerwerth@unibw.de}
\end{minipage}
\vspace{0.1cm}

\hspace{-0.6cm}Julius Beerwerth received the B.Sc. degree in 2020 and M.Sc. degree in 2022, both in Computational Engineering Science from RWTH Aachen University, Germany. He is currently pursuing the Ph.D. degree in Control Engineering at the Department of Aerospace Engineering, University of the Bundeswehr Munich. His research interests include data-driven control for cyber-physical systems and its applications to connected and automated vehicles.

\vspace{0.5cm}

\hspace{-0.7cm}
\begin{minipage}{0.3\columnwidth}
\includegraphics[width=1in,height=1.25in,clip,keepaspectratio]{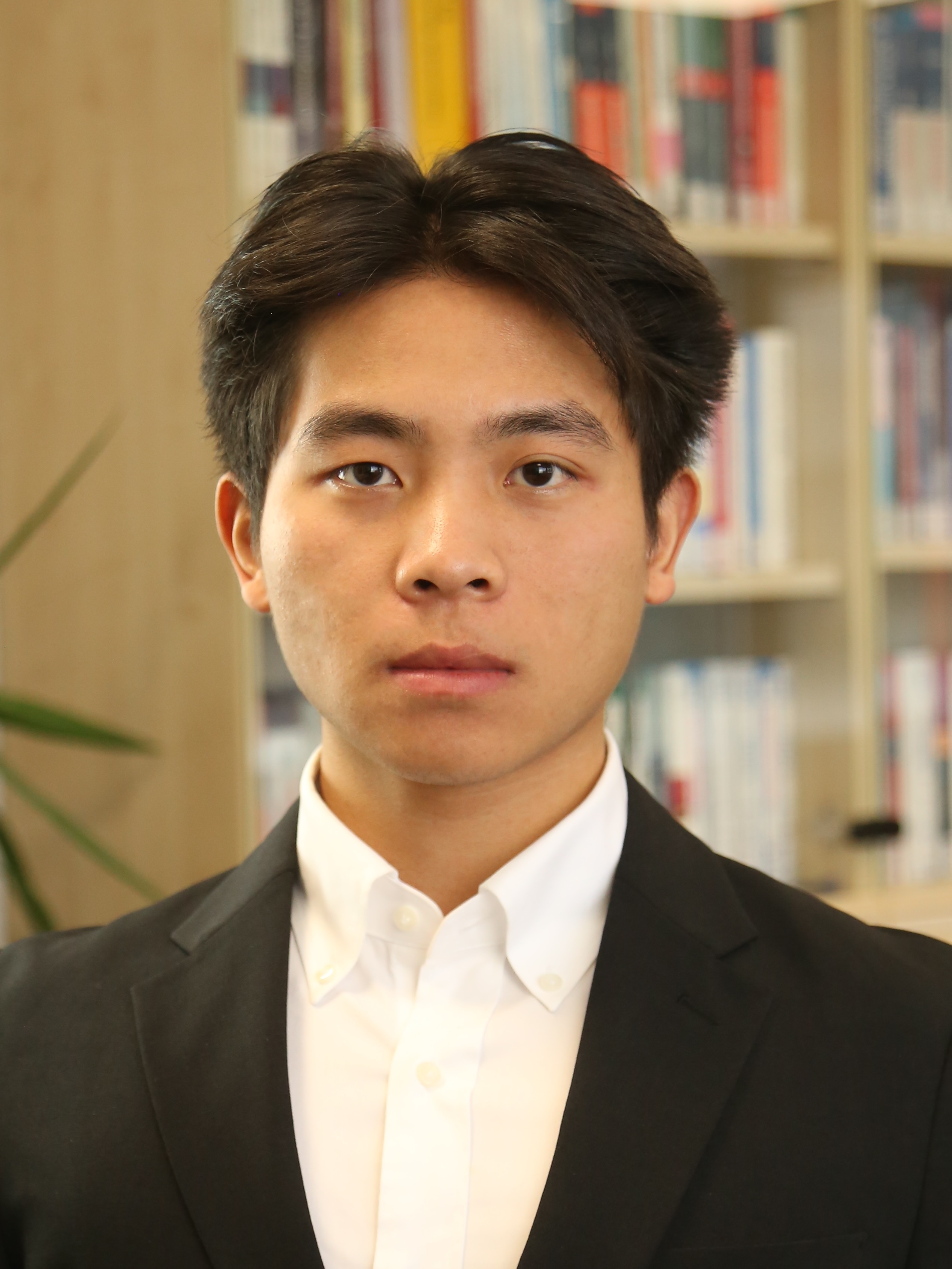}
\end{minipage}
\begin{minipage}{0.7\columnwidth}
\textbf{Jianye Xu}  \\
Dept. of Computer Science, \\RWTH Aachen University, \\
52074 Aachen, \\
Germany \\
\textbf{xu@embedded.rwth-aachen.de}
\end{minipage}
\vspace{0.1cm}

\hspace{-0.6cm}Jianye Xu received a B.Sc. degree in Mechanical Engineering from the Beijing Institute of Technology, China, in 2020, and an M.Sc. degree in Automation Engineering from RWTH Aachen University, Germany, in 2022. He is currently pursuing a Ph.D. in Computer Science at RWTH Aachen University. His research focuses on learning and optimization-based multi-agent decision-making and its applications in connected and automated vehicles.

\vspace{0.5cm}

\hspace{-0.7cm}
\begin{minipage}{0.3\columnwidth}
\includegraphics[width=1in,height=1.25in,clip,keepaspectratio]{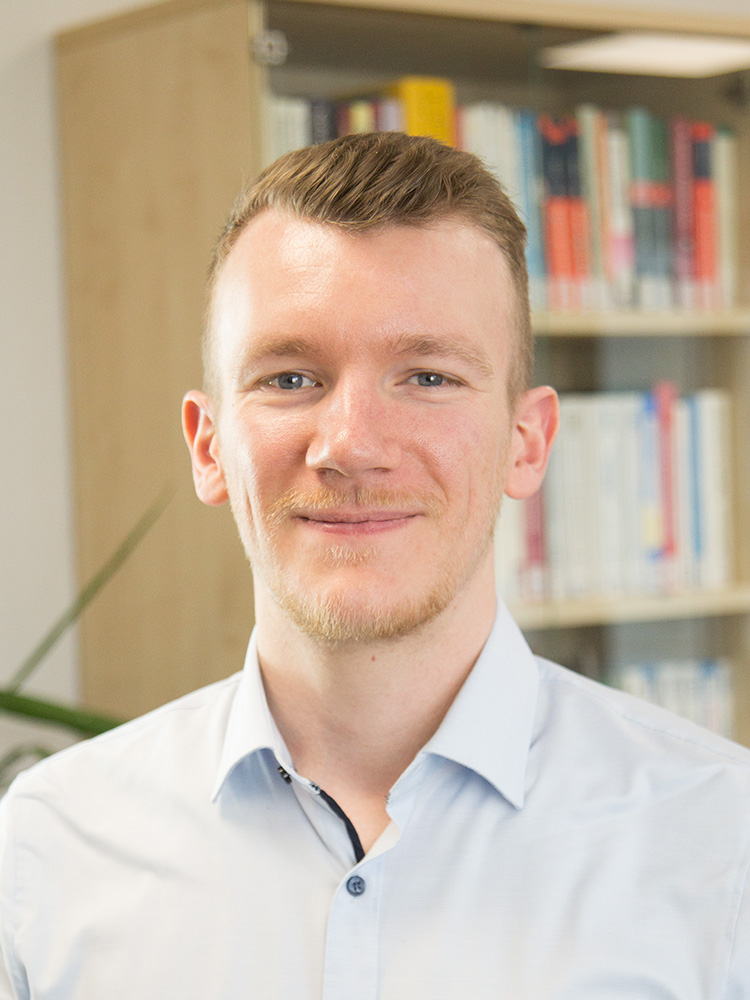}
\end{minipage}
\begin{minipage}{0.7\columnwidth}
\textbf{Simon Schäfer}  \\
Dept. of Computer Science, \\RWTH Aachen University, \\
52074 Aachen, \\
Germany \\
\textbf{schaefer@embedded.rwth-aachen.de}
\end{minipage}
\vspace{0.1cm}

\hspace{-0.6cm}Simon Schäfer received the B.Sc. degree in 2019 and M.Sc. degree in 2021, both in Computational Engineering Science from RWTH Aachen University, Germany. He is currently pursuing the Ph.D. degree in vehicle localization at the Department of Computer Science, RWTH Aachen University. His research interests include infrastructure-based vehicle localization and software deployment for cyber-physical systems.

\vspace{0.5cm}

\hspace{-0.7cm}
\begin{minipage}{0.3\columnwidth}
\includegraphics[width=1in,height=1.25in,clip,keepaspectratio]{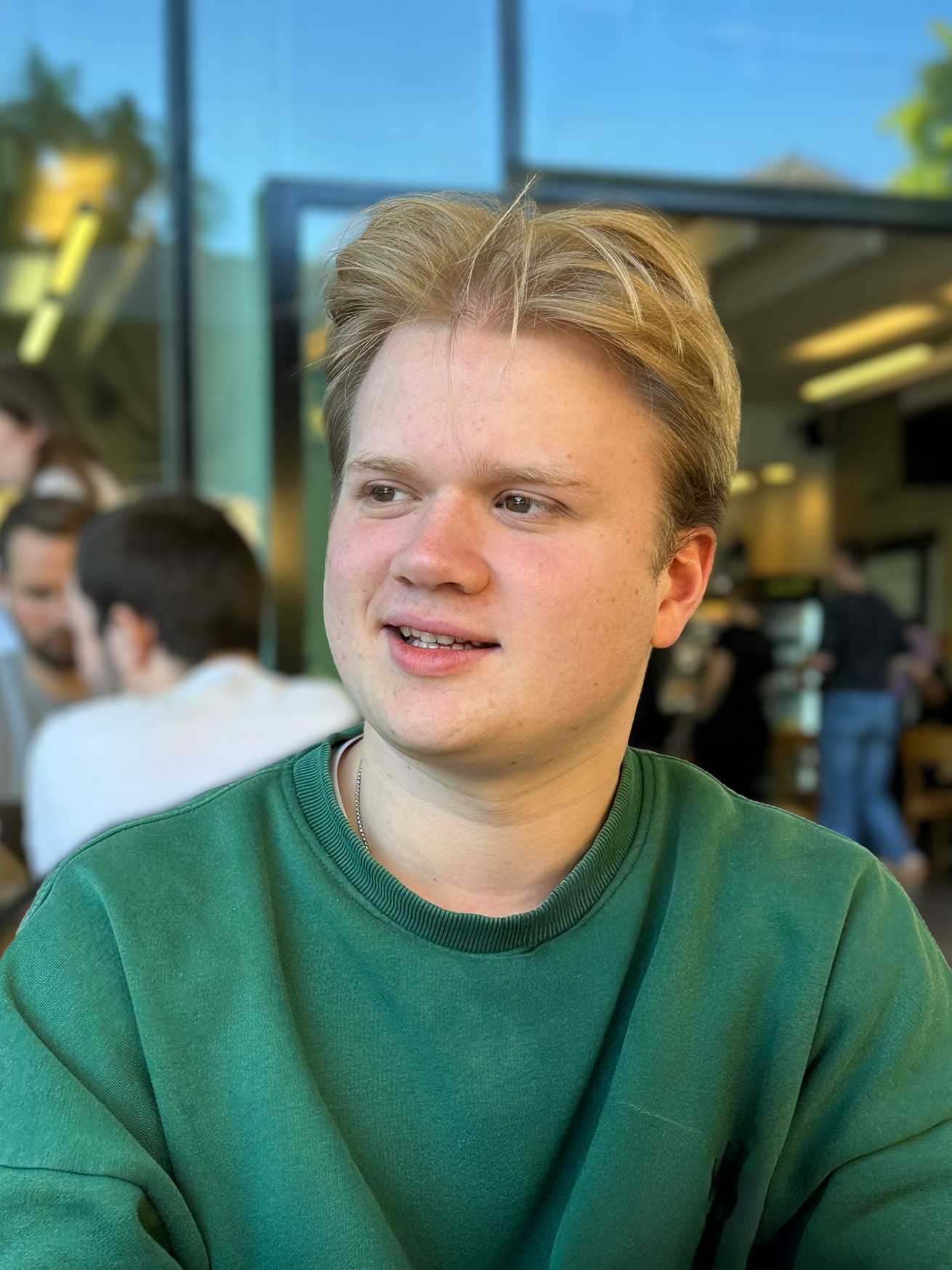}
\end{minipage}
\begin{minipage}{0.7\columnwidth}
\textbf{Fynn Belderink}  \\
Dept. of Computer Science, \\RWTH Aachen University, \\
52074 Aachen, \\
Germany \\
\textbf{belderink@embedded.rwth-aachen.de}
\end{minipage}
\vspace{0.1cm}

\hspace{-0.6cm}Fynn Belderink is currently pursuing his undergraduate degree in Computer Science at RWTH Aachen University, Germany. He is currently a student assistant at the Department of Computer Science, RWTH Aachen University. His research interests include planning in multi-agent reinforcement learning and autonomous driving.

\vspace{0.5cm}

\hspace{-0.7cm}
\begin{minipage}{0.3\columnwidth}
\includegraphics[width=1in,height=1.25in,clip,keepaspectratio]{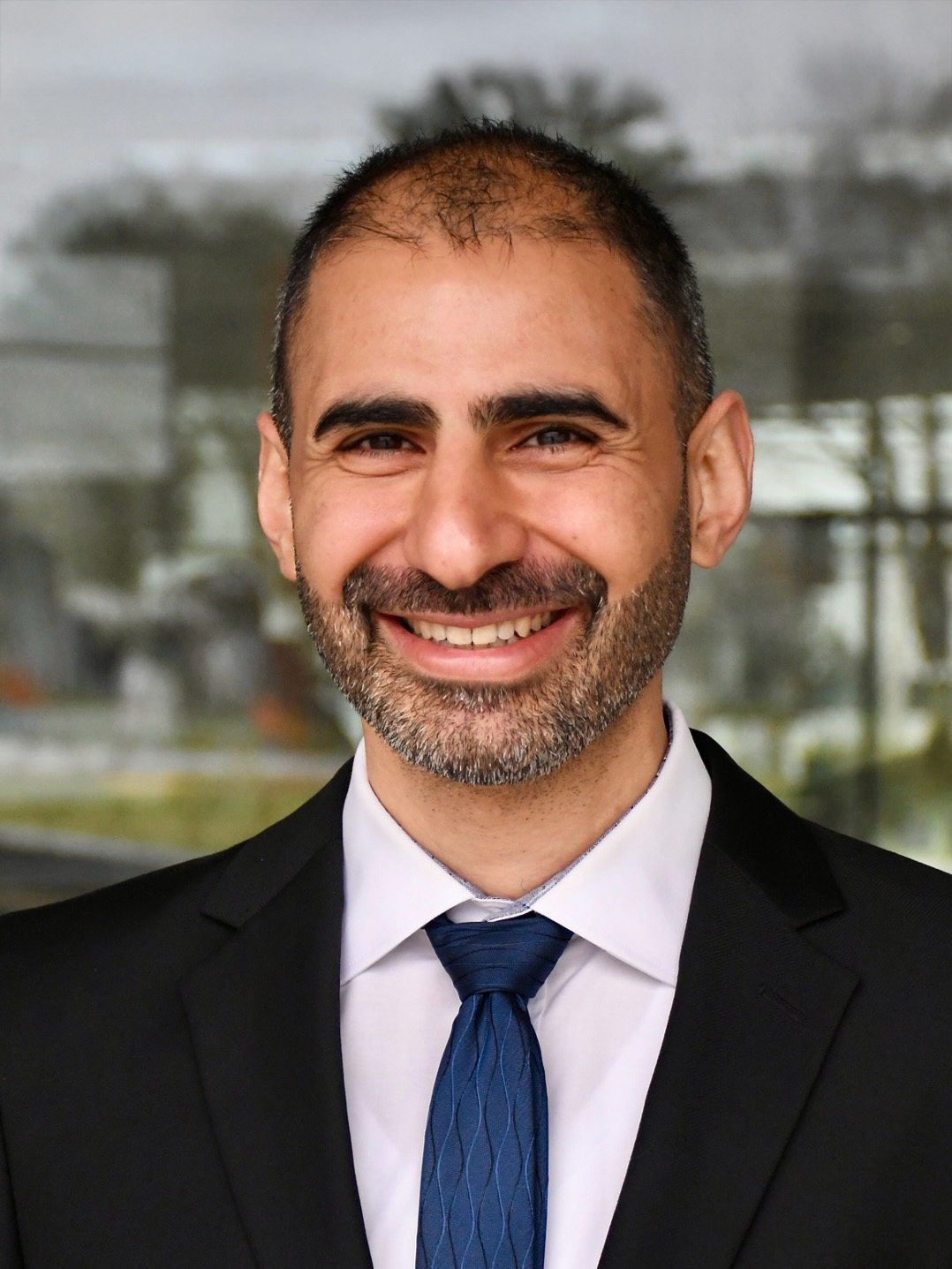}
\end{minipage}
\begin{minipage}{0.7\columnwidth}
\textbf{Bassam Alrifaee}  \\
Dept. of Aerospace Engineering, \\University of the Bundeswehr Munich, \\
85579 Neubiberg, \\
Germany \\
\textbf{bassam.alrifaee@unibw.de}
\end{minipage}
\vspace{0.1cm}

\hspace{-0.6cm}Prof. Bassam Alrifaee holds the Professorship for Adaptive Behavior of Autonomous Vehicles in the Department of Aerospace Engineering at the University of the Bundeswehr (UniBw) Munich. His research focuses on the intelligent control of autonomous systems, with particular emphasis on distributed control, cooperative localization, software architectures, and experimental validation.
Before joining UniBw Munich in 2024, he served as a Senior Researcher and Lecturer at RWTH Aachen University, where he founded the Cyber-Physical Mobility (CPM) group and the CPM Lab. In 2023, he was a Visiting Scholar at the Information and Decision Science Laboratory at the University of Delaware, USA.
Prof. Alrifaee has secured research grants from various institutions and received awards for his advisory and editorial contributions. He is a Senior Member of the IEEE. 
\end{document}

%% file: 00_acronyms.tex
\DeclareAcronym{cav}{
    short = CAV,
    long  = Connected and Automated Vehicle,
}
\DeclareAcronym{cbf}{
    short = CBF,
    long  = Control Barrier Function,
}
\DeclareAcronym{cg}{
    short = CG,
    long = Center of Gravity,
    short-plural = s,
    long-plural-form = Centers of Gravity,
}
\DeclareAcronym{cnn}{
    short = CNN,
    long  = Convolutional Neural Network
}
\DeclareAcronym{cpm}{
    short = CPM,
    long  = Cyber-Physical Mobility
}
\DeclareAcronym{cpmlab}{
    short = CPM Lab,
    long  = \ac{cpm} Lab
}
\DeclareAcronym{dmpc}{
    short = DMPC,
    long  = distributed model predictive control
}
\DeclareAcronym{dql}{
    short = DQL,
    long  = Deep Q-Learning
}

\DeclareAcronym{hd}{
    short = HD,
    long  = High-Definition
}
\DeclareAcronym{il}{
    short = IL,
    long  = Imitation Learning,
    short-indefinite = an,
}
\DeclareAcronym{mappo}{
    short = MAPPO,
    long  = Multi-Agent \ac{ppo},
    short-indefinite = an,
}
\DeclareAcronym{maddpg}{
    short = MADDPG,
    long  = Multi-Agent Deep Deterministic Policy Gradient,
    short-indefinite = an,
}
\DeclareAcronym{mas}{
    short = MAS,
    long  = Multi-Agent System,
    short-indefinite = an,
}
\DeclareAcronym{mdp}{
    short = MDP,
    long  = Markov decision process,
    short-indefinite = an,
}
\DeclareAcronym{mg}{
    short = MG,
    long  = Markov Game,
    short-indefinite = an,
}
\DeclareAcronym{mpc}{
    short = MPC,
    long  = model predictive control,
    short-indefinite = an,
}
\DeclareAcronym{marl}{
    short = MARL,
    long  = Multi-Agent Reinforcement Learning,
    short-indefinite = an,
}
\DeclareAcronym{ocp}{
    short = OCP,
    long  = Optimal Control Problem,
    short-indefinite = an,
    long-indefinite = an,
}
\DeclareAcronym{per}{
    short = PER,
    long  = Prioritized Experience Replay
}
\DeclareAcronym{pomdp}{
    short = POMDP,
    long  = Partially Observable \ac{mdp}
}
\DeclareAcronym{pomg}{
    short = POMG,
    long  = Partially Observable \ac{mg}
}
\DeclareAcronym{ppo}{
    short = PPO,
    long  = Proximal Policy Optimization
}
\DeclareAcronym{rhc}{
    short = RHC,
    long  = receding horizon control,
    short-indefinite = an,
}
\DeclareAcronym{rl}{
    short = RL,
    long  = Reinforcement Learning,
    short-indefinite = an,
}
\DeclareAcronym{zsg}{
    short = ZSG,
    long  = Zero-Shot Generalization,
}

%% file: figures/prediction-illustration.tex
{
\scalefont{1.23456}

\begin{tikzpicture}[scale=0.81,transform shape]
    \tikzstyle{label}=[anchor=north, outer sep = 0.1cm]
    \tikzstyle{text_below}=[outer sep=0.03cm, inner sep=0cm, minimum height=0.6cm]

    \tikzstyle{field_arrow}=[->]

    \draw[black, dashed] (2cm,-2.5cm) -- (2cm,0.3cm);
    \draw[black, dashed] (2cm,3.2cm) -- (2cm,1.3cm);

    \node[label, fill=white] at (2cm,0cm) {$t$};

    \node[label] at (3cm,0cm) {$t+1$};

    \draw[black, dashed] (7cm,-2.5cm) -- (7cm,2.3cm);
    \node[label, fill=white] at (7cm,0cm) {$t+H_c$};
  
    \draw[black, dashed] (10cm,-2.5cm) -- (10cm,2.3cm);
    \node[label, fill=white] at (10cm,0cm) {$t+H_p$};
    
    \draw[field_arrow] (2cm,2.8cm) -- (1cm,2.8cm);
    \node[text_below, anchor=south east] at (1.8cm,2.8cm) {Past};

    \draw[field_arrow] (2cm,2.8cm) -- (3cm,2.8cm);
    \node[text_below, anchor=south west] at (2.2cm,2.8cm) {Future};
    
    \draw[field_arrow] (2cm,-1.6cm) -- (7cm,-1.6cm);
    \node[text_below, anchor=south west, fill=white] at (2.2cm,-1.6cm) {Policy-based $\bm{u}_{{t+j}}^{(i)}$};

    \draw[field_arrow] (7cm,-1.6cm) -- (10cm,-1.6cm);
    \node[text_below, anchor=south west, fill=white] at (7.2cm,-1.6cm) {Rules-based $\bm{u}_{{t+j}}^{(i)}$};

    \draw[field_arrow] (2cm,-2.4cm) -- (10cm,-2.4cm);
    \node[text_below, anchor=south west, fill=white] at (2.2cm,-2.4cm) {Predicted states $\bm{s}_{t,\mathrm{SigmaRL}}^{(i)}$};

\begin{scope}[yshift=-2.8cm]

    \draw[field_arrow] (0.5cm,4cm) -- (10.5cm,4cm);
    \draw[field_arrow] (2cm,3.9cm) -- (2cm,5cm);
    \node[label, anchor=west] at (2cm,5cm) {$u_v$};

    \draw[black] (1cm,3.9cm) -- (1cm,4.1cm);
    \draw[black] (2cm,3.9cm) -- (2cm,4.1cm);
    \draw[black] (3cm,3.9cm) -- (3cm,4.1cm);
    \draw[black] (4cm,3.9cm) -- (4cm,4.1cm);
    \draw[black] (5cm,3.9cm) -- (5cm,4.1cm);
    \draw[black] (6cm,3.9cm) -- (6cm,4.1cm);
    \draw[black] (7cm,3.9cm) -- (7cm,4.1cm);
    \draw[black] (8cm,3.9cm) -- (8cm,4.1cm);
    \draw[black] (9cm,3.9cm) -- (9cm,4.1cm);
    \draw[black] (10cm,3.9cm) -- (10cm,4.1cm);

    \draw[black] (1cm, 4.7cm) -- (2cm, 4.7cm);

    \draw[black, line width=0.05cm] (2cm, 4.7cm) -- (2cm, 4.5cm) -- (3cm, 4.5cm) -- (3cm, 4.8cm) -- (4cm, 4.8cm) -- (4cm, 4.3cm) -- (5cm, 4.3cm) -- (5cm, 4.85cm) -- (6cm, 4.85cm) -- (6cm, 4.4cm) -- (7cm, 4.4cm) -- (7cm, 4.6cm) -- (8cm, 4.6cm) -- (8cm, 4.6cm) -- (9cm, 4.6cm) -- (9cm, 4.6cm) -- (10cm, 4.6cm);
\end{scope}
\begin{scope}[yshift=-3.0cm]

    \draw[field_arrow] (0.5cm,3cm) -- (10.5cm,3cm);
    \draw[field_arrow] (2cm,2.9cm) -- (2cm,4cm);
    \node[label, anchor=west] at (2cm,4cm) {$|u_\sigma|$};

    \draw[black] (1cm,2.9cm) -- (1cm,3.1cm);
    \draw[black] (2cm,2.9cm) -- (2cm,3.1cm);
    \draw[black] (3cm,2.9cm) -- (3cm,3.1cm);
    \draw[black] (3cm,2.9cm) -- (3cm,3.1cm);
    \draw[black] (5cm,2.9cm) -- (5cm,3.1cm);
    \draw[black] (6cm,2.9cm) -- (6cm,3.1cm);
    \draw[black] (7cm,2.9cm) -- (7cm,3.1cm);
    \draw[black] (8cm,2.9cm) -- (8cm,3.1cm);
    \draw[black] (9cm,2.9cm) -- (9cm,3.1cm);
    \draw[black] (10cm,2.9cm) -- (10cm,3.1cm);

    \draw[black] (1cm, 3.7cm) -- (2cm, 3.7cm);

    \draw[black, line width=0.05cm] (2cm, 3.7cm) -- (2cm, 3.5cm) -- (3cm, 3.5cm) -- (3cm, 3.1cm) -- (4cm, 3.1cm) -- (4cm, 3.8cm) -- (5cm, 3.8cm) -- (5cm, 3.5cm) -- (6cm, 3.5cm) -- (6cm, 3.6cm) -- (7cm, 3.6cm) -- (7cm, 3.4cm) -- (8cm, 3.4cm) -- (8cm, 3.2cm) -- (9cm, 3.2cm) -- (9cm, 3.0cm) -- (10cm, 3.0cm);


\end{scope}

\end{tikzpicture}
}

%% file: figures/experiment-overview.tex
\begin{subfigure}[b]{7cm}
    \centering
    \includegraphics[width=7cm]{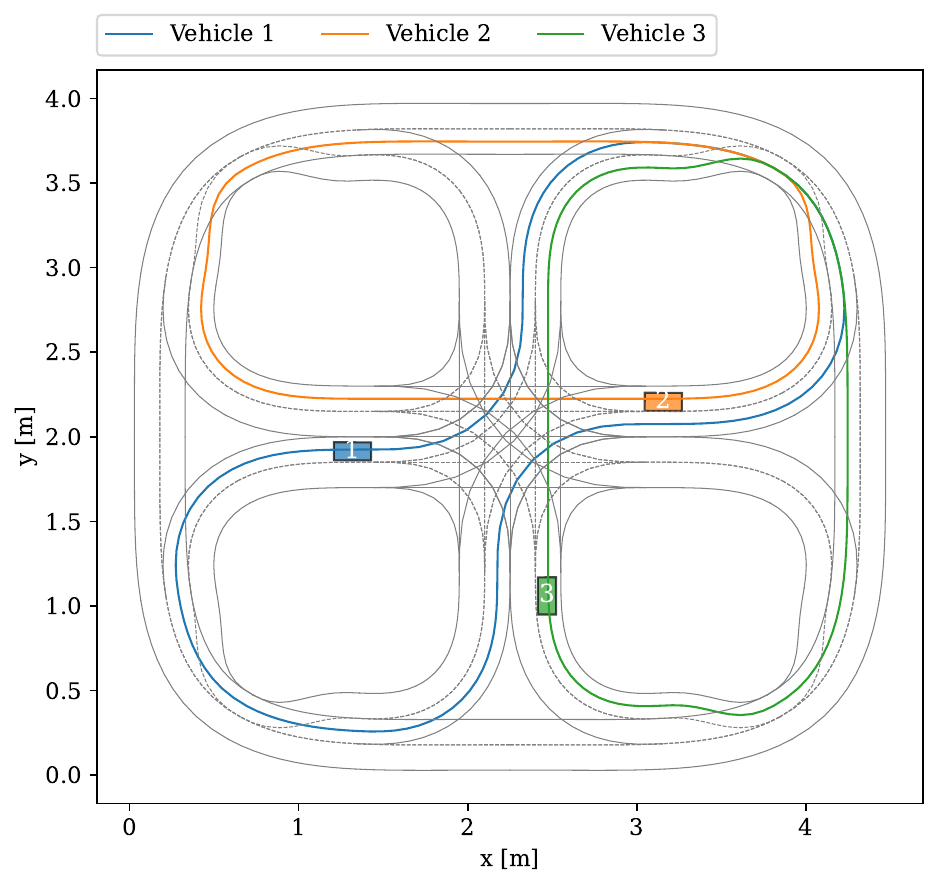}
    \caption{Initial setup.}
\end{subfigure}
\begin{subfigure}[b]{7cm}
    \centering
    \includegraphics[width=7cm]{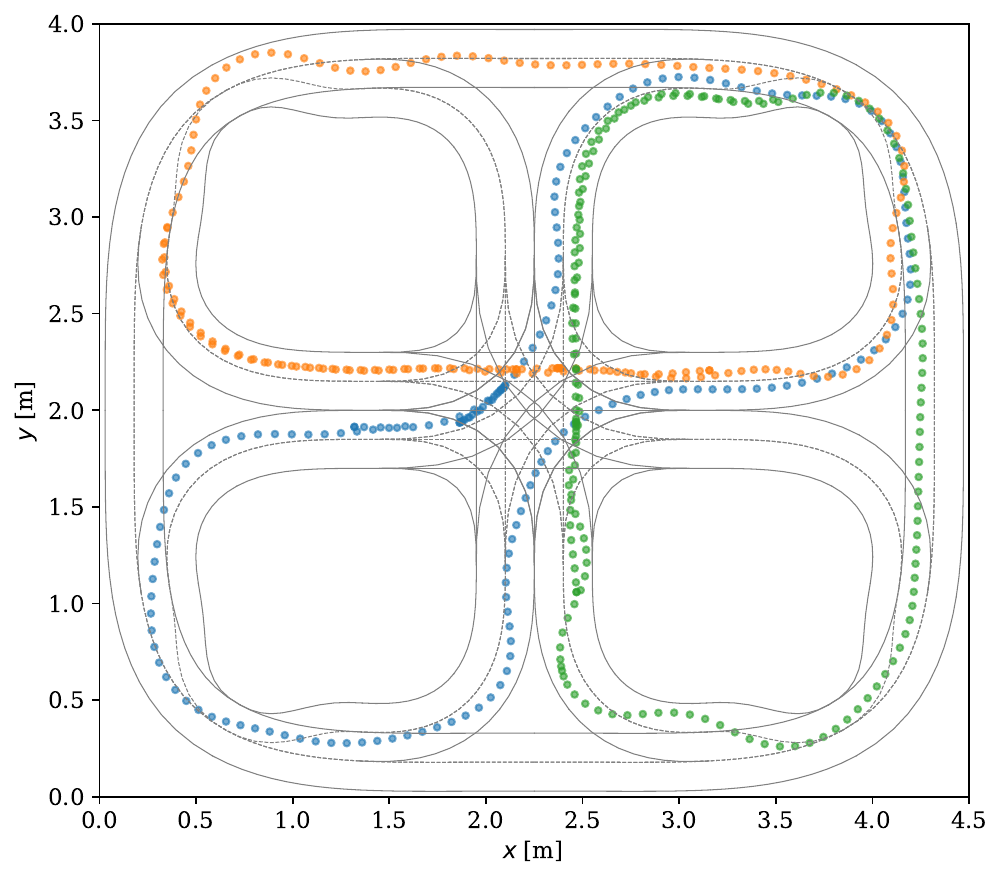}
    \caption{SigmaRL Simulation.}
\end{subfigure}
\begin{subfigure}[b]{7cm}
    \centering
    \includegraphics[width=7cm]{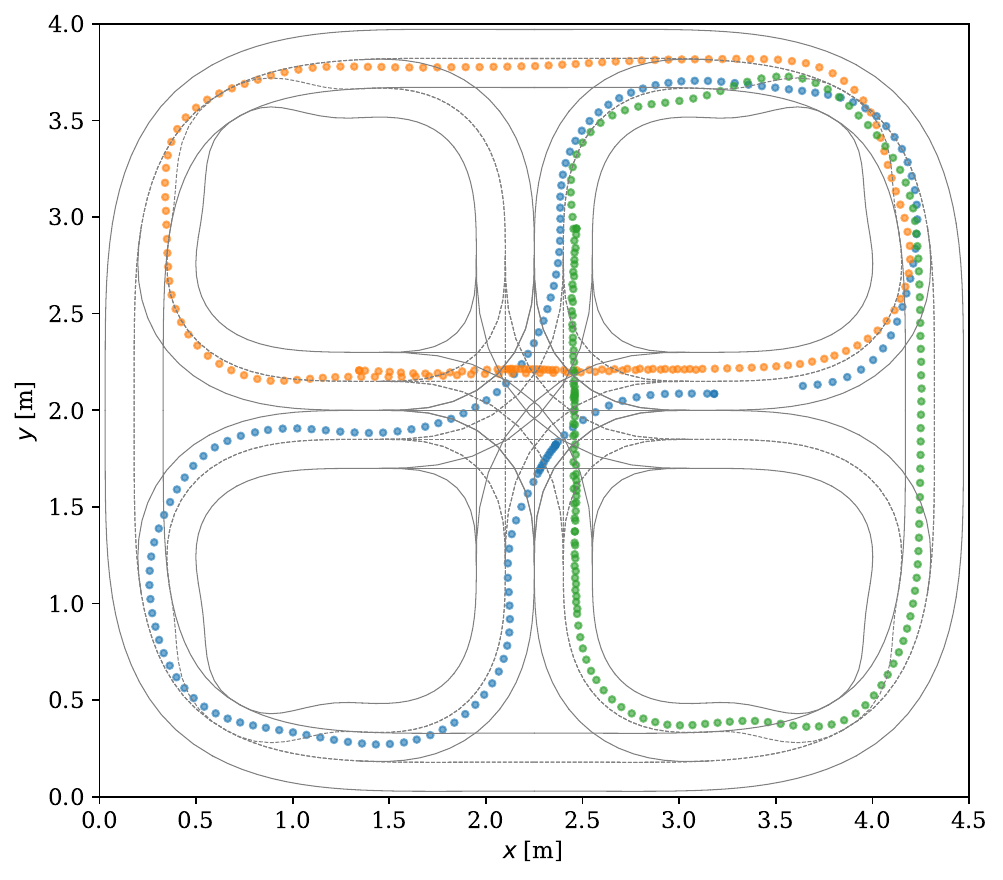}
    \caption{CPM Lab Digital Twin.}
\end{subfigure}
\begin{subfigure}[b]{7cm}
    \centering
    \includegraphics[width=7cm]{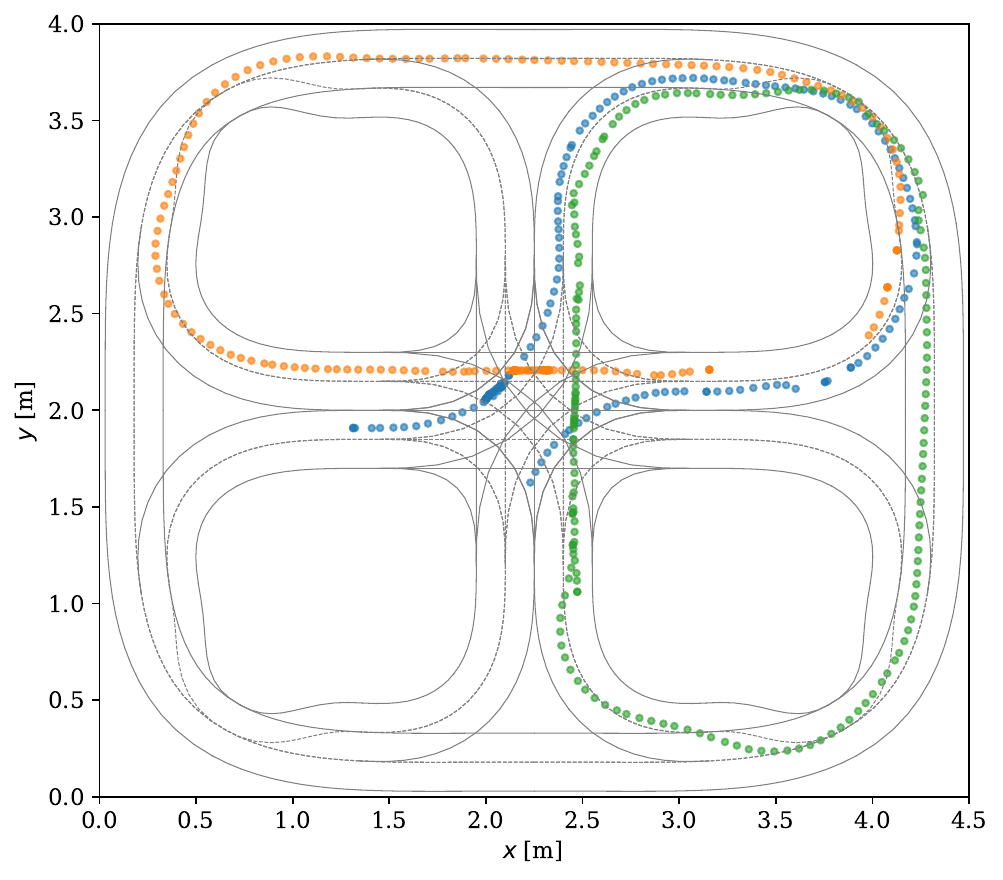}
    \caption{CPM Lab Physical Experiment.}
\end{subfigure}